\documentclass{article}

\usepackage[preprint]{neurips_2022}
\usepackage{graphicx}
\usepackage[colorlinks=true,linkcolor=blue, citecolor=blue]{hyperref}%

\usepackage[utf8]{inputenc} 
\usepackage[T1]{fontenc}    
\usepackage{hyperref}       
\usepackage{url}            
\usepackage{booktabs}       
\usepackage{amsfonts}       
\usepackage{nicefrac}       
\usepackage{microtype}      
\usepackage{xcolor}         

\title{Zero-Shot and Few-Shot Learning for Lung Cancer Multi-Label Classification using Vision Transformer}

\author{%
  Fu-Ming Guo\thanks{fmguo@acm.org, Association for Computing Machinery.} \\
  \And
   Yingfang Fan\thanks{Harvard Medical School.} \\
}

\begin{document}

\maketitle

\begin{abstract}
Lung cancer is the leading cause of cancer-related death worldwide. Lung adenocarcinoma (LUAD) and lung squamous cell carcinoma (LUSC) are the most common histologic subtypes of non-small-cell lung cancer (NSCLC). Histology is an essential tool for lung cancer diagnosis. Pathologists make classifications according to the dominant subtypes. Although morphology remains the standard for diagnosis, significant tool needs to be developed to elucidate the diagnosis. In our study, we utilize the pre-trained Vision Transformer (ViT) model to classify multiple label lung cancer on histologic slices (from dataset LC25000), in both Zero-Shot and Few-Shot settings. Then we compare the performance of Zero-Shot and Few-Shot ViT on accuracy, precision, recall, sensitivity and specificity. Our study show that the pre-trained ViT model has a good performance in Zero-Shot setting, a competitive accuracy ($99.87\%$) in Few-Shot setting ({epoch = 1}) and an optimal result ($100.00\%$ on both validation set and test set) in Few-Shot seeting ({epoch = 5}).

\end{abstract}

\section{Introduction}

Lung cancer is the leading cause of cancer-related death worldwide. It is not only because of smoking, but also exposure to toxic chemicals. Non-small-cell lung cancer (NSCLC) is any malignant epithelial lung tumor that lacks a small-cell component (\cite{mengoli20182015}). NSCLC represents approximately $85\%$ of all new lung cancer diagnoses (\cite{gridelli2015non}). Lung adenocarcinoma (LUAD) and lung squamous cell carcinoma (LUSC) are the most common histologic subtypes (\cite{herbst2018biology}) of NSCLC. These subtypes are further subclassified into multiple subtypes according to WHO criteria (\cite{travis20152015}). Histology is an essential factor for individualizing treatment based on either safety or efficacy outcomes (\cite{langer2010evolving}). 
Adenocarcinomas are malignant epithelial tumors with glandular differentiation. It has clear morphologic patterns such as acinar, papillary, lepidic, micropapillary (\cite{travis20152015}), although mix pattern adenocarcinomas are most common (\cite{pmid30955514}). Squamous cell carcinomas are often centrally located and derived from bronchial epithelial cells. Unequivocal keratinization and well-formed classical bridges can be diagnosed as squamous cell carcinomas (\cite{travis20152015}).

Histologic distinctions may be unclear due to poorly differentiated tumors and requires confirmatory immunohistochemical stains. The heterogeneous histology within the same lesion occurs in many NSCLC tumors. Pathologists make classifications according to the dominant subtypes. Although morphology remains the standard for diagnosis, significant tool needs to be developed to elucidate the diagnosis.

Using AI to analyze tissue sections is typically called computational pathology (\cite{fuchs2011computational}). Research in this area can trace back to the middle of the last century, with the seminal application of image analysis algorithms to medical images. Image analysis algorithms can classify cell images based on quantitative cell characteristics, e.g., size, shape, and chromatin distribution, and support the diagnosis of diseases (\cite{mendelsohn1965computer}).The early applications implemented computational features matched to a biological process, later replaced by radionics using generic features of texture descriptors (\cite{zwanenburg2020image}).

Automated classification of abnormal lesions using images is a challenging task owing to the fine-grained variability in the appearance of abnormal lesions. Deep convolutional neural networks (CNN) (\cite{lecun1999object}) show potential for general and highly variable tasks across many fine-grained object categories. In the recent booming (\cite{dean2022golden}) of Deep Learning(\cite{lecun2015deep}), a series of CNN-based models unceasingly refreshed the state-of-the-art performance on various computer vision benchmarks (\cite{deng2009imagenet}; \cite{krizhevsky2009learning}). Nowadays, the Transformer (\cite{vaswani2017attention}) shows an advantage over computer vision tasks (\cite{dosovitskiy2020image}) after becoming the de-facto module for natural language processing tasks (\cite{devlin2019bert}; \cite{brown2020language}).

\begin{figure*}[ht]
    \centering
    \includegraphics[width=13.5cm]{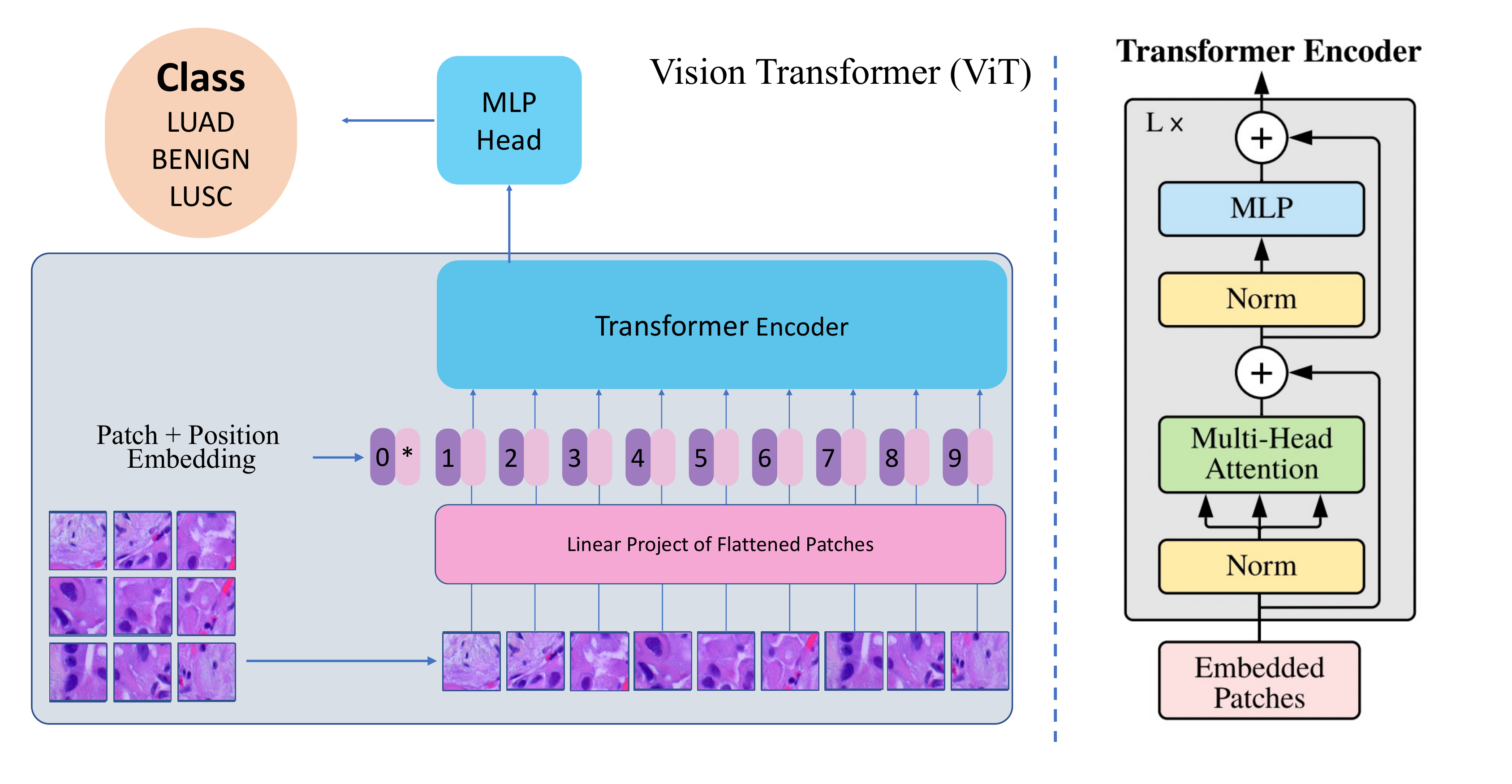}
    \caption{Model Overview - Vision Transformer for Lung Cancer Classification}
    \label{fig:model_overview}
\end{figure*}

The main advantage of the Deep Learning approach is automatically learning features from the data, instead of crafting meaningful features in feature engineering and conventional image analysis. The automated feature learning from the Deep Learning approach reduced the required domain knowledge and the implementation time. More importantly, the automated Deep Learning approach yields robust, hierarchical feature representations, which outperform traditional image analysis methods in most cases. 

Despite the powerful learning ability of the Deep Learning approach, there is still the central issue in the application in the medical domain, the domain shift. How do these state-of-the-art computer vision models perform on medical image analysis tasks? Specifically, the histological analysis of lung cancer in our case. The Zero-Shot Learning (\cite{larochelle2008zero};\cite{socher2013zero};\cite{wei2021finetuned}) and Few-Shot Learning (\cite{brown2020language}) are the emerging paradigms to address this issue. Few-shot transfer learning, where a model is first pre-trained on a data-rich task before being fine-tuned on a wide range of downstream tasks, has emerged as a powerful technique in natural languages and computer vision tasks. Pre-trained  representations have shown substantial performance improvements using self-supervised learning and transfer learning. There are also emerging zero-shot learning techniques showing outstanding performance, like prompt-tuning (\cite{li-liang-2021-prefix}) and instruction tuning (\cite{wei2021finetuned}). In our study, we utilize the pre-trained Vision Transformer (ViT from \cite{dosovitskiy2020image}) model to classify multiple label lung cancer on histologic slices (from dataset LC25000 in  \cite{borkowski2019lung}), in both Zero-Shot and Few-Shot manners. Then we compare the performance of Zero-Shot and Few-Shot ViT on accuracy, precision, recall, sensitivity and specificity. Our study show that the pre-trained ViT model has a good performance in Zero-Shot setting, a competitive accuracy in Few-Shot setting ({epoch = 1}) and an optimal result in Few-Shot seeting ({epoch = 5}).

To illustrate the working mechanism behind these models, we utilized the cutting-edge model interpretation work Transformer models (\cite{chefer2021transformer}) compared with the analysis from the human expert. Our pathology expert provides the information about which regions help her decide whether the input image is a lung cancer lesion or not from the professional perspective. The comparison between the pathology cognition and Grad-CAM visualization presents a high degree of consistency. This comparison helps us understand our deep learning model, and rationalizes our proposed method further.

\section{Methods}
\label{headings}

The pre-train-finetune paradigm is the robust defacto pipeline in visual and language learning. In the finetune stage, a Multi-Layer Perceptron (MLP) (\cite{rosenblatt1958perceptron}; 
\cite{aizerman1964theoretical}) often follows a thoroughly pre-trained backbone network to function as a projector and aid transfer learning (\cite{wang2021revisiting}) on new tasks (medical image classification in our case). Our customized models utilized the paradigm above: ViT as the pre-trained backbone, followed by an MLP projector, and finally, a cross-entropy loss to classify and diagnose the cancer lesion (Figure \ref{fig:model_overview}). 

\subsection{Model Architectures}
We utilize the ViT (\cite{dosovitskiy2020image}) as our Transformer based backbone. The ViT backbone is pretrained on ImageNet (\cite{deng2009imagenet}), ImageNet-21k (\cite{ridnik2021imagenetk}), and JFT-300M (\cite{sun2017revisiting}). ViT backbones holds accuracy of {88.05\%} on ImageNet (\cite{deng2009imagenet}), {94.55\%} on CIFAR-100 (\cite{krizhevsky2009learning}) and {77.63\%} on the VTAB (\cite{zhai2019visual}) suite of 19 tasks. In Figure \ref{fig:model_overview}, we illustrate how the ViT backbone function in our customize d model. The multiple scale input image samples ({768 * 768} pixels) are first normalized to {224 * 224} pixels. This follows the input image format in ImageNet, due to ViT is pre-trained on ImageNet. For the 224 x 224 pixels input, the ViT backbone first split the input image into 196 image patches, and each patch is 16 x 16 pixels. Image patches are treated the same way as tokens (words) in natural language processing applications. The image patches are linearly embedded, with positional embeddings. ViT backbone (\cite{dosovitskiy2020image}) then feeds the resulting sequence vectors into a standard Transformer encoder (\cite{vaswani2017attention}). The process is visually illustrated in Figure \ref{fig:model_overview}.

\section{Experiments}
\label{others}

In this section, we describe the dataset we used and our experiments in the Zero-Shot and Few-Shot manners.

\subsection{Dataset}

We utilized the Lung Cancer part of the dataset LC25000 (\cite{borkowski2019lung}). LC25000 has 25,000 color images in 5 classes. Each class contains 5,000 images of the histologic categories: colon adenocarcinoma, benign colonic tissue, lung adenocarcinoma, lung squamous cell carcinoma and benign lung tissue. All images are de-identified, HIPPA\footnote{Health Insurance Portability and Accountability Act of 1996 (HIPAA), https://www.cdc.gov/phlp/publications/topic/hipaa.html} compliant and validated. We utilize the lung cancer categories: lung adenocarcinoma, lung squamous cell carcinoma and benign lung tissue. We split the 15,000 images in lung cancer categories into training set (${D_{train}}$), validation set (${D_{validation}}$) and test set (${D_{test}}$), by the ratios ${60\%, 20\%, 20\%}$, after random sampling.

\subsection{Zero-Shot transfer learning}

First, we conducted the experiments in a Zero-Shot manner. The well pre-trained ViT model functions as frozen wights in this setting. We directly use the frozen pre-trained ViT model to make the prediction on test set (${D_{test}}$). We report the accuracy on validation set and test set in Table \ref{table1}.

\subsection{Few-Shot transfer learning}

Then we conducted the experiments in Few-Shot manners. We fine-tuned the pre-trained ViT model on training set (${D_{train}}$) for 5 epochs, and validation the best model on validation set (${D_{validation}}$) at each epoch. We find that the pre-trained ViT has prompt and strong learning ability, so that it quick achieves the optimal accuracy after fine-tuning of only 5 epochs ($100.00\%$ on both validation set (${D_{validation}}$) and test set (${D_{test}}$)). We report the accuracies on validation set and test set in Table \ref{table1}.

\begin{table}
  \caption{Zero-Shot and Few-Shot performance of ViT on LC 25000}
  \label{table1}
  \centering
  \begin{tabular}{lll}
    \toprule
    \multicolumn{2}{c}{Part}                   \\
    \cmidrule(r){1-2}
    Name     & Validation set (acc)     & Test set (acc) \\
    \midrule
    Zero-Shot &   & $33.77\%$     \\
    Few-Shot (${epoch = 1}$)     & $98.90\%$ & $98.87\%$      \\
    Few-Shot (${epoch = 2}$)     & $98.47\%$ & $99.50\%$      \\
    Few-Shot (${epoch = 3}$)     & $99.77\%$ & $99.70\%$      \\
    Few-Shot (${epoch = 4}$)     & $98.90\%$ & $99.87\%$      \\
    Few-Shot (${epoch = 5}$)     & $100.00\%$ & $100.00\%$      \\
    \bottomrule
  \end{tabular}
\end{table}

\subsection{Receiver operating characteristic (ROC) curve}

\begin{figure*}[ht]
    \centering
    \includegraphics[width=13.5cm]{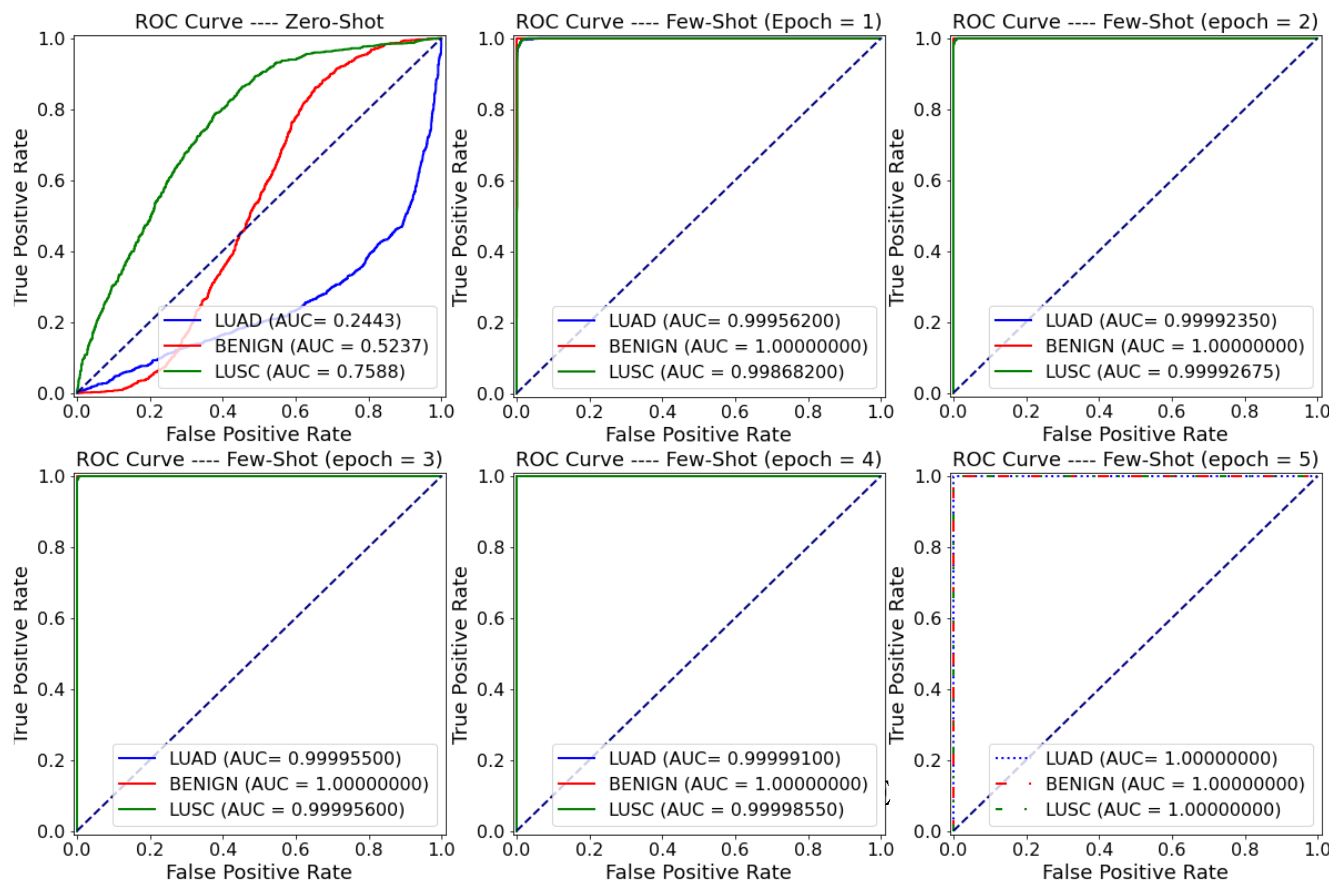}
    \caption{Receiver operating characteristic (ROC) curves for Zero-Shot and Few-Shot ViT models}
    \label{fig:roc}
\end{figure*}

The Receiver operating characteristic (ROC) curve is used to evaluate the quality of a classifier (\cite{green1966signal}; \cite{fawcett2006introduction}). The x axis of ROC curve is the ‘false positive rate’ (FPR), and the y axis means the ‘true positive rate’ (TPR). The point or line of a classifier locates more top-left on the ROC curve, the better this classifier is. Area under the Curve of ROC (AUC ROC), tests whether positives are ranked higher than negatives. We reported the AUC values of each model (of all the three classes, LUAD, BENIGH and LUSC) on Figure \ref{fig:roc}. The AUC values of ViT model (Few-Shot $epoch = 5$) is 1.00000000 for all the three classes (LUAD, BENIGN abd LUSC), showing that the ViT model (Few-Shot $epoch = 5$) is an optimal classifier in our case.

\section{Discussion}

\begin{figure*}[ht]
    \centering
    \includegraphics[width=13.5cm]{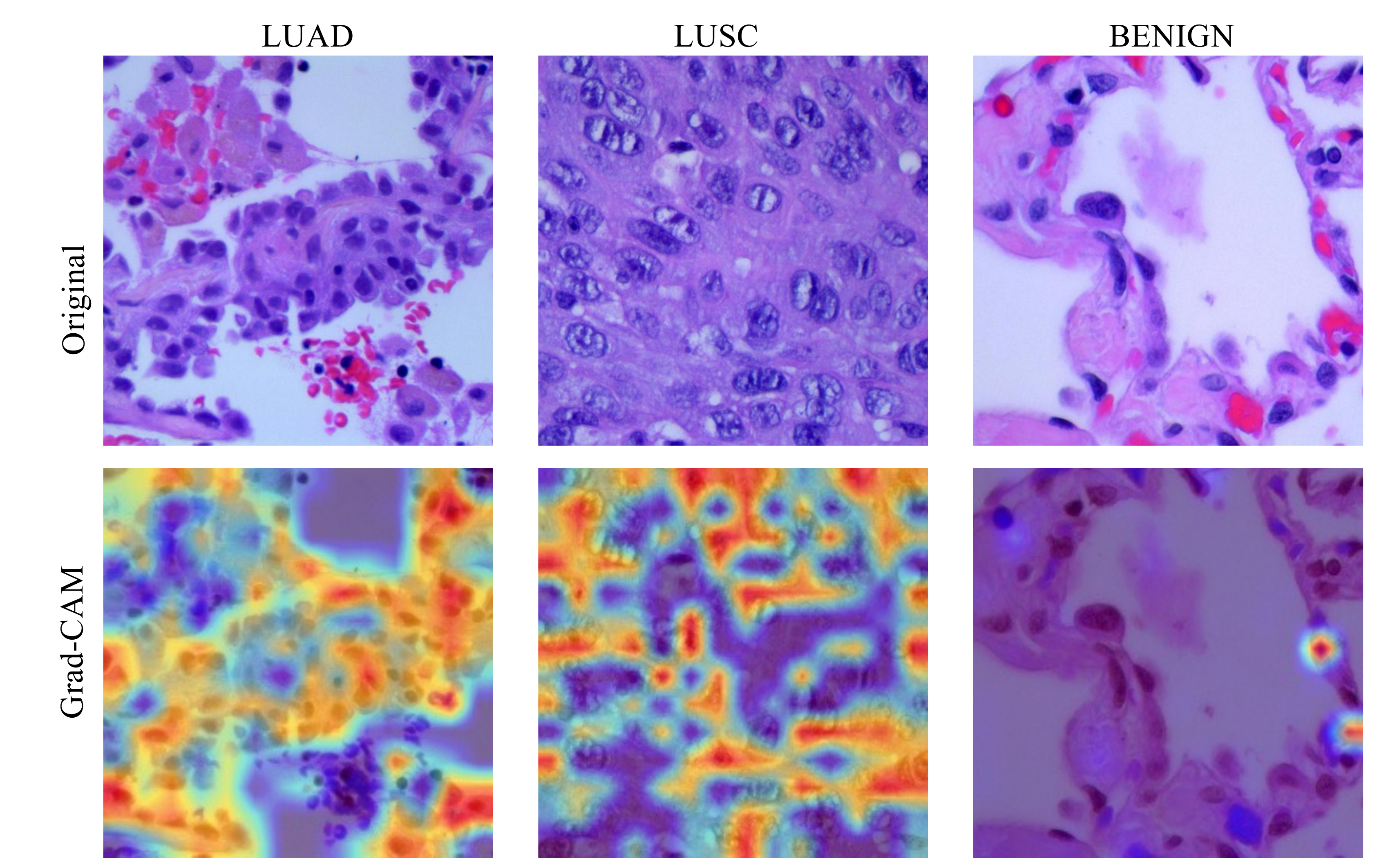}
    \caption{Attention Visualization: first row -- Original histological image; 
second row -- Grad-CAM -- discriminative regions.  }
    \label{fig:vis}
\end{figure*}

A lot of work endeavors to make deep learning more sensible and explainable. In various deep learning applications especially into medical imaging, it is crucial to make the deep learning model more interpretative. \cite{selvaraju2017grad} have introduced a Gradient Weighted Class Activation Mapping (Grad-CAM) technique which provides the interpretative view of deep learning models. (\cite{chefer2021transformer}) extends the research of Grad-CAM to Transformer \cite{vaswani2017attention} models. Grad-CAM uses the gradients of any target concept, flowing into the final convolutional layer to produce the coarse localization map highlighting important regions in the image for predicting the concept. To illustrate the working mechanism behind these models, we utilized the cutting-edge model interpretation work Transformer models (\cite{chefer2021transformer}) compared with the analysis from the human expert. In our application to classify the histology images, our visualizations (Figure \ref{fig:vis}) lend insights into failure modes of these models, showing that seemingly unreasonable predictions have reasonable explanations. Our visualization (Figure \ref{fig:vis}) is robust to adversarial perturbations, are more faithful to the underlying model and help achieve model generalizations by identifying dataset bias.
The interpretation work of deep learning models is essential to the understanding of the mechanism behind the success of our model, and making the model more transparent. In our case, we utilize Grad-CAM to generate the visualized explanations via gradient-based localization. The localizations in Grad-CAM help explain which regions (confirmed salient features) in the image input contribute more to the model's final decision and which regions are less (Figure \ref{fig:vis}). Furthermore, our pathology expert provides the information about which regions help him decide whether the input image is an infection or not from the professional perspective. The comparison between the pathology cognition and Grad-CAM visualization presents a high degree of consistency. This comparison is novel, helps us understand our deep learning model, and rationalizes our proposed method further.

\begin{ack}
We thank the computing resource support from the TPU Research Cloud. Both Guo FM and Fan Y contributed to manuscript revision, read, and approved the submitted version.
\end{ack}

\bibliography{neurips}

\begin{thebibliography}{33}
\providecommand{\natexlab}[1]{#1}
\providecommand{\url}[1]{\texttt{#1}}
\expandafter\ifx\csname urlstyle\endcsname\relax
  \providecommand{\doi}[1]{doi: #1}\else
  \providecommand{\doi}{doi: \begingroup \urlstyle{rm}\Url}\fi

\bibitem[Aizerman(1964)]{aizerman1964theoretical}
Mark~A Aizerman.
\newblock Theoretical foundations of the potential function method in pattern
  recognition learning.
\newblock \emph{Automation and remote control}, 25:\penalty0 821--837, 1964.

\bibitem[Borkowski et~al.(2019)Borkowski, Bui, Thomas, Wilson, DeLand, and
  Mastorides]{borkowski2019lung}
Andrew~A Borkowski, Marilyn~M Bui, L~Brannon Thomas, Catherine~P Wilson,
  Lauren~A DeLand, and Stephen~M Mastorides.
\newblock Lung and colon cancer histopathological image dataset (lc25000).
\newblock \emph{arXiv preprint arXiv:1912.12142}, 2019.

\bibitem[Brown et~al.(2020)Brown, Mann, Ryder, Subbiah, Kaplan, Dhariwal,
  Neelakantan, Shyam, Sastry, Askell, et~al.]{brown2020language}
Tom~B Brown, Benjamin Mann, Nick Ryder, Melanie Subbiah, Jared Kaplan, Prafulla
  Dhariwal, Arvind Neelakantan, Pranav Shyam, Girish Sastry, Amanda Askell,
  et~al.
\newblock Language models are few-shot learners.
\newblock \emph{arXiv preprint arXiv:2005.14165}, 2020.

\bibitem[Chefer et~al.(2021)Chefer, Gur, and Wolf]{chefer2021transformer}
Hila Chefer, Shir Gur, and Lior Wolf.
\newblock Transformer interpretability beyond attention visualization.
\newblock In \emph{Proceedings of the IEEE/CVF Conference on Computer Vision
  and Pattern Recognition}, pages 782--791, 2021.

\bibitem[Dean(2022)]{dean2022golden}
Jeffrey Dean.
\newblock A golden decade of deep learning: Computing systems \& applications.
\newblock \emph{Daedalus}, 151\penalty0 (2):\penalty0 58--74, 2022.

\bibitem[Deng et~al.(2009)Deng, Dong, Socher, Li, Li, and
  Fei-Fei]{deng2009imagenet}
Jia Deng, Wei Dong, Richard Socher, Li-Jia Li, Kai Li, and Li~Fei-Fei.
\newblock Imagenet: A large-scale hierarchical image database.
\newblock In \emph{2009 IEEE conference on computer vision and pattern
  recognition}, pages 248--255. Ieee, 2009.

\bibitem[Devlin et~al.(2019)Devlin, Chang, Lee, and Toutanova]{devlin2019bert}
Jacob Devlin, Ming-Wei Chang, Kenton Lee, and Kristina Toutanova.
\newblock Bert: Pre-training of deep bidirectional transformers for language
  understanding.
\newblock In \emph{Proceedings of the 2019 Conference of the North American
  Chapter of the Association for Computational Linguistics: Human Language
  Technologies, Volume 1 (Long and Short Papers)}, pages 4171--4186, 2019.

\bibitem[Dosovitskiy et~al.(2020)Dosovitskiy, Beyer, Kolesnikov, Weissenborn,
  Zhai, Unterthiner, Dehghani, Minderer, Heigold, Gelly,
  et~al.]{dosovitskiy2020image}
Alexey Dosovitskiy, Lucas Beyer, Alexander Kolesnikov, Dirk Weissenborn,
  Xiaohua Zhai, Thomas Unterthiner, Mostafa Dehghani, Matthias Minderer, Georg
  Heigold, Sylvain Gelly, et~al.
\newblock An image is worth 16x16 words: Transformers for image recognition at
  scale.
\newblock \emph{arXiv preprint arXiv:2010.11929}, 2020.

\bibitem[Fawcett(2006)]{fawcett2006introduction}
Tom Fawcett.
\newblock An introduction to roc analysis.
\newblock \emph{Pattern recognition letters}, 27\penalty0 (8):\penalty0
  861--874, 2006.

\bibitem[Fuchs and Buhmann(2011)]{fuchs2011computational}
Thomas~J Fuchs and Joachim~M Buhmann.
\newblock Computational pathology: challenges and promises for tissue analysis.
\newblock \emph{Computerized Medical Imaging and Graphics}, 35\penalty0
  (7-8):\penalty0 515--530, 2011.

\bibitem[Green et~al.(1966)Green, Swets, et~al.]{green1966signal}
David~Marvin Green, John~A Swets, et~al.
\newblock \emph{Signal detection theory and psychophysics}, volume~1.
\newblock Wiley New York, 1966.

\bibitem[Gridelli et~al.(2015)Gridelli, Rossi, Carbone, Guarize, Karachaliou,
  Mok, Petrella, Spaggiari, and Rosell]{gridelli2015non}
Cesare Gridelli, Antonio Rossi, David~P Carbone, Juliana Guarize, Niki
  Karachaliou, Tony Mok, Francesco Petrella, Lorenzo Spaggiari, and Rafael
  Rosell.
\newblock Non-small-cell lung cancer.
\newblock \emph{Nature reviews Disease primers}, 1\penalty0 (1):\penalty0
  1--16, 2015.

\bibitem[Herbst et~al.(2018)Herbst, Morgensztern, and
  Boshoff]{herbst2018biology}
Roy~S Herbst, Daniel Morgensztern, and Chris Boshoff.
\newblock The biology and management of non-small cell lung cancer.
\newblock \emph{Nature}, 553\penalty0 (7689):\penalty0 446--454, 2018.

\bibitem[Krizhevsky et~al.(2009)Krizhevsky, Hinton,
  et~al.]{krizhevsky2009learning}
Alex Krizhevsky, Geoffrey Hinton, et~al.
\newblock Learning multiple layers of features from tiny images.
\newblock 2009.

\bibitem[Langer et~al.(2010)Langer, Besse, Gualberto, Brambilla, and
  Soria]{langer2010evolving}
Corey~J Langer, Benjamin Besse, Antonio Gualberto, Elizabeth Brambilla, and
  Jean-Charles Soria.
\newblock The evolving role of histology in the management of advanced
  non--small-cell lung cancer.
\newblock \emph{Journal of clinical oncology}, 28\penalty0 (36):\penalty0
  5311--5320, 2010.

\bibitem[Larochelle et~al.(2008)Larochelle, Erhan, and
  Bengio]{larochelle2008zero}
Hugo Larochelle, Dumitru Erhan, and Yoshua Bengio.
\newblock Zero-data learning of new tasks.
\newblock In \emph{AAAI}, volume~1, page~3, 2008.

\bibitem[LeCun et~al.(1999)LeCun, Haffner, Bottou, and Bengio]{lecun1999object}
Yann LeCun, Patrick Haffner, L{\'e}on Bottou, and Yoshua Bengio.
\newblock Object recognition with gradient-based learning.
\newblock In \emph{Shape, contour and grouping in computer vision}, pages
  319--345. Springer, 1999.

\bibitem[LeCun et~al.(2015)LeCun, Bengio, and Hinton]{lecun2015deep}
Yann LeCun, Yoshua Bengio, and Geoffrey Hinton.
\newblock Deep learning.
\newblock \emph{nature}, 521\penalty0 (7553):\penalty0 436--444, 2015.

\bibitem[Li and Liang(2021)]{li-liang-2021-prefix}
Xiang~Lisa Li and Percy Liang.
\newblock Prefix-tuning: Optimizing continuous prompts for generation.
\newblock In \emph{Proceedings of the 59th Annual Meeting of the Association
  for Computational Linguistics and the 11th International Joint Conference on
  Natural Language Processing (Volume 1: Long Papers)}, pages 4582--4597,
  Online, August 2021. Association for Computational Linguistics.
\newblock \doi{10.18653/v1/2021.acl-long.353}.

\bibitem[Mendelsohn et~al.(1965)Mendelsohn, Kolman, Perry, and
  Prewitt]{mendelsohn1965computer}
Mortimer~L Mendelsohn, Wilfred~A Kolman, Benson Perry, and Judith~MS Prewitt.
\newblock Computer analysis of cell images.
\newblock \emph{Postgraduate Medicine}, 38\penalty0 (5):\penalty0 567--573,
  1965.

\bibitem[Mengoli et~al.(2018)Mengoli, Longo, Fraggetta, Cavazza, Dubini, Ali,
  Guddo, Gilioli, Bogina, Nannini, et~al.]{mengoli20182015}
MC~Mengoli, FR~Longo, F~Fraggetta, A~Cavazza, A~Dubini, G~Ali, F~Guddo,
  E~Gilioli, G~Bogina, N~Nannini, et~al.
\newblock The 2015 world health organization classification of lung tumors: new
  entities since the 2004 classification.
\newblock \emph{Pathologica-Journal of the Italian Society of Anatomic
  Pathology and Diagnostic Cytopathology}, 110\penalty0 (1):\penalty0 39--67,
  2018.

\bibitem[Nasim et~al.(2019)Nasim, Sabath, and Eapen]{pmid30955514}
F.~Nasim, B.~F. Sabath, and G.~A. Eapen.
\newblock {{L}ung {C}ancer}.
\newblock \emph{Med Clin North Am}, 103\penalty0 (3):\penalty0 463--473, May
  2019.

\bibitem[Ridnik et~al.(2021)Ridnik, Ben-Baruch, Noy, and
  Zelnik-Manor]{ridnik2021imagenetk}
Tal Ridnik, Emanuel Ben-Baruch, Asaf Noy, and Lihi Zelnik-Manor.
\newblock Imagenet-21k pretraining for the masses.
\newblock In \emph{Thirty-fifth Conference on Neural Information Processing
  Systems Datasets and Benchmarks Track (Round 1)}, 2021.
\newblock URL \url{https://openreview.net/forum?id=Zkj_VcZ6ol}.

\bibitem[Rosenblatt(1958)]{rosenblatt1958perceptron}
Frank Rosenblatt.
\newblock The perceptron: a probabilistic model for information storage and
  organization in the brain.
\newblock \emph{Psychological review}, 65\penalty0 (6):\penalty0 386, 1958.

\bibitem[Selvaraju et~al.(2017)Selvaraju, Cogswell, Das, Vedantam, Parikh, and
  Batra]{selvaraju2017grad}
Ramprasaath~R Selvaraju, Michael Cogswell, Abhishek Das, Ramakrishna Vedantam,
  Devi Parikh, and Dhruv Batra.
\newblock Grad-cam: Visual explanations from deep networks via gradient-based
  localization.
\newblock In \emph{Proceedings of the IEEE international conference on computer
  vision}, pages 618--626, 2017.

\bibitem[Socher et~al.(2013)Socher, Ganjoo, Manning, and Ng]{socher2013zero}
Richard Socher, Milind Ganjoo, Christopher~D Manning, and Andrew Ng.
\newblock Zero-shot learning through cross-modal transfer.
\newblock \emph{Advances in neural information processing systems}, 26, 2013.

\bibitem[Sun et~al.(2017)Sun, Shrivastava, Singh, and Gupta]{sun2017revisiting}
Chen Sun, Abhinav Shrivastava, Saurabh Singh, and Abhinav Gupta.
\newblock Revisiting unreasonable effectiveness of data in deep learning era.
\newblock In \emph{Proceedings of the IEEE international conference on computer
  vision}, pages 843--852, 2017.

\bibitem[Travis et~al.(2015)Travis, Brambilla, Nicholson, Yatabe, Austin,
  Beasley, Chirieac, Dacic, Duhig, Flieder, et~al.]{travis20152015}
William~D Travis, Elisabeth Brambilla, Andrew~G Nicholson, Yasushi Yatabe,
  John~HM Austin, Mary~Beth Beasley, Lucian~R Chirieac, Sanja Dacic, Edwina
  Duhig, Douglas~B Flieder, et~al.
\newblock The 2015 world health organization classification of lung tumors:
  impact of genetic, clinical and radiologic advances since the 2004
  classification.
\newblock \emph{Journal of thoracic oncology}, 10\penalty0 (9):\penalty0
  1243--1260, 2015.

\bibitem[Vaswani et~al.(2017)Vaswani, Shazeer, Parmar, Uszkoreit, Jones, Gomez,
  Kaiser, and Polosukhin]{vaswani2017attention}
Ashish Vaswani, Noam Shazeer, Niki Parmar, Jakob Uszkoreit, Llion Jones,
  Aidan~N Gomez, Lukasz Kaiser, and Illia Polosukhin.
\newblock Attention is all you need.
\newblock In \emph{NIPS}, 2017.

\bibitem[Wang et~al.(2021)Wang, Tang, Zhu, Bai, Zhao, Qi, and
  Ouyang]{wang2021revisiting}
Yizhou Wang, Shixiang Tang, Feng Zhu, Lei Bai, Rui Zhao, Donglian Qi, and Wanli
  Ouyang.
\newblock Revisiting the transferability of supervised pretraining: an mlp
  perspective.
\newblock \emph{arXiv preprint arXiv:2112.00496}, 2021.

\bibitem[Wei et~al.(2021)Wei, Bosma, Zhao, Guu, Yu, Lester, Du, Dai, and
  Le]{wei2021finetuned}
Jason Wei, Maarten Bosma, Vincent~Y Zhao, Kelvin Guu, Adams~Wei Yu, Brian
  Lester, Nan Du, Andrew~M Dai, and Quoc~V Le.
\newblock Finetuned language models are zero-shot learners.
\newblock \emph{arXiv preprint arXiv:2109.01652}, 2021.

\bibitem[Zhai et~al.(2019)Zhai, Puigcerver, Kolesnikov, Ruyssen, Riquelme,
  Lucic, Djolonga, Pinto, Neumann, Dosovitskiy, et~al.]{zhai2019visual}
Xiaohua Zhai, Joan Puigcerver, Alexander Kolesnikov, Pierre Ruyssen, Carlos
  Riquelme, Mario Lucic, Josip Djolonga, Andre~Susano Pinto, Maxim Neumann,
  Alexey Dosovitskiy, et~al.
\newblock The visual task adaptation benchmark.
\newblock 2019.

\bibitem[Zwanenburg et~al.(2020)Zwanenburg, Valli{\`e}res, Abdalah, Aerts,
  Andrearczyk, Apte, Ashrafinia, Bakas, Beukinga, Boellaard,
  et~al.]{zwanenburg2020image}
Alex Zwanenburg, Martin Valli{\`e}res, Mahmoud~A Abdalah, Hugo~JWL Aerts,
  Vincent Andrearczyk, Aditya Apte, Saeed Ashrafinia, Spyridon Bakas, Roelof~J
  Beukinga, Ronald Boellaard, et~al.
\newblock The image biomarker standardization initiative: standardized
  quantitative radiomics for high-throughput image-based phenotyping.
\newblock \emph{Radiology}, 295\penalty0 (2):\penalty0 328--338, 2020.

\end{thebibliography}
\medskip


\end{document}